\documentclass[runningheads]{llncs}
\usepackage{graphicx}
\usepackage{siunitx}
\usepackage{cite}
\usepackage{amssymb}
\usepackage[utf8]{inputenc}
\usepackage{tabularx}
\usepackage{verbatim}
\usepackage{booktabs}
\usepackage{array}
\begin{document}
\title{Learning to See Forces: Surgical Force Prediction with RGB-Point Cloud Temporal Convolutional Networks}
\titlerunning{Learning to See Forces}
\author{Cong Gao\and
Xingtong Liu\and
Michael Peven\and
Mathias Unberath\and
Austin Reiter}
\authorrunning{Cong Gao et al.}
\institute{The Johns Hopkins University, Baltimore MD 21218, USA\\
\email{cgao11@jhu.edu}\\}
\maketitle              % typeset the header of the contribution
\begin{abstract}
Robotic surgery has been proven to offer clear advantages during surgical procedures, however, one of the major limitations is obtaining haptic feedback. Since it is often challenging to devise a hardware solution with accurate force feedback, we propose the use of ``visual cues'' to infer forces from tissue deformation. Endoscopic video is a passive sensor that is freely available, in the sense that any minimally-invasive procedure already utilizes it. To this end, we employ deep learning to infer forces from video as an attractive low-cost and accurate alternative to typically complex and expensive hardware solutions. First, we demonstrate our approach in a phantom setting using the da Vinci Surgical System affixed with an OptoForce sensor. Second, we then validate our method on an \textit{ex vivo} liver organ. Our method results in a mean absolute error of \SI{0.814}{\newton} in the \textit{ex vivo} study, suggesting that it may be a promising alternative to hardware based surgical force feedback in endoscopic procedures.

%\keywords{First keyword  \and Second keyword \and Another keyword.}
\end{abstract}
%
%
% Introduction
%
\section{Introduction}
Robot-assisted clinical systems have been increasingly adopted due to their advantages during surgical procedures. However, obtaining haptic feedback of a teleoperated surgical system still constitutes a hard problem due to practical challenges such as control loop stability. In the current version of the da Vinci Surgical System~\cite{ref_article1} (Intuitive Surgical, Inc., Sunnyvale, CA, USA), there is no haptic technology and no feedback on the grip forces. Surgeons depend on visual cues to infer the forces to avoid damage to tools and anatomy since excessive mechanical force can lead to the breakage of an end-effector string, serious artery or nerve injury, and even post-operation trauma~\cite{ref_article2}. As a result, there is a critical need to design force sensing systems in the field of surgical robotics.

Recently, many researchers have focused their efforts on solutions to this problem. For instance, numerous tactile sensing devices have been developed to estimate tactile information during static (point based) measurements, including indentation-based contact devices, aspiration devices, optical fiber devices, and non-contact devices~\cite{ref_article3}. Such devices are capable of providing accurate tactile information during static measurements of a single point, but they cannot scan soft tissue in a dynamic way, which is not a real-time solution~\cite{ref_article4}.  Another area of investigation is directed towards using a torque sensor to model and compensate for grip force. This may provide a consistent internal force compensation based on the quantitative model, but it largely relies on the surgeon's skills and experience~\cite{ref_article2}. In addition, most of these hardware-based solutions have delicate and expensive components, which often cannot withstand sterilization. 

Vision based approaches are one way to overcome above limitations of hardware solutions. Starting from~\cite{ref_article5}, computer vision has been used to measure the deformed object and recover the applied force from linear elasticity equations. Recent advances in deep learning bring opportunities to such vision based force prediction in real surgical scenarios~\cite{ref_article6, ref_article7, ref_article8, ref_article9}. For example, researchers in [6] extract the 3D deformable structure of the heart and use a neural network with the architecture of LSTM-RNN to predict the applied force. 

\begin{figure}[h]
\includegraphics[width=\textwidth]{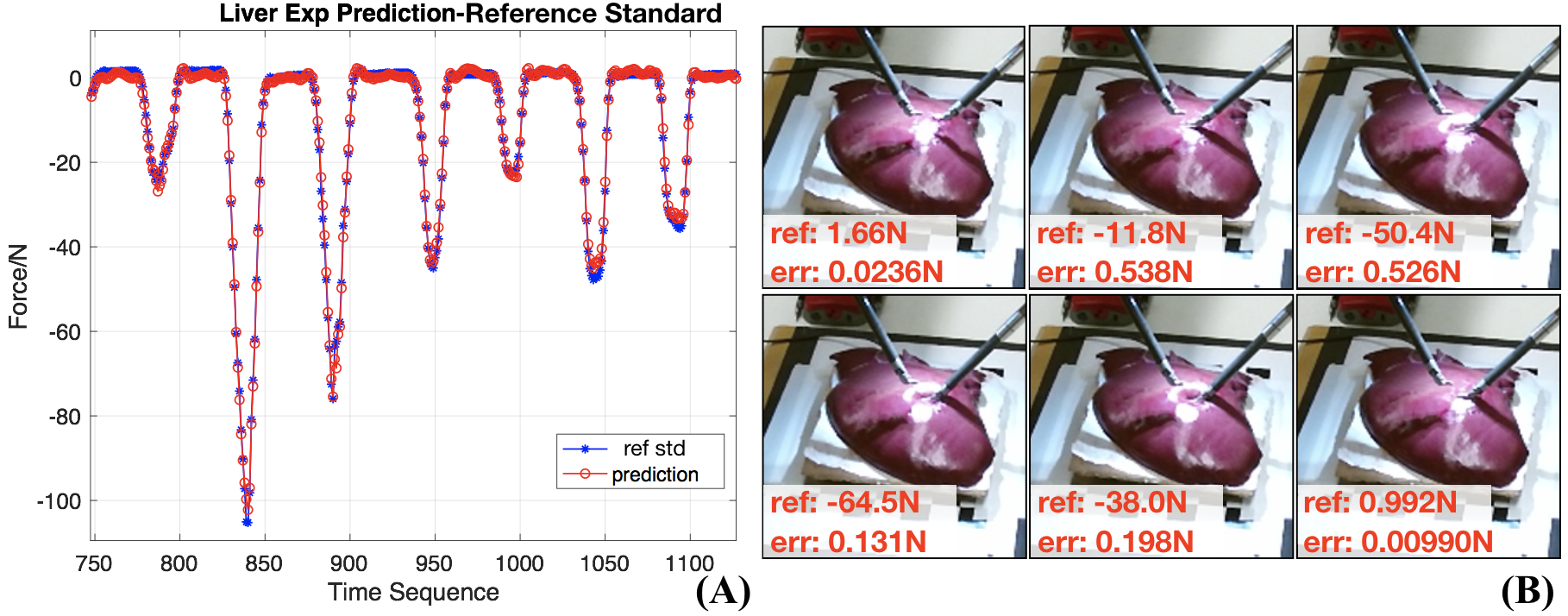}
\caption{Results of the \textit{ex vivo} liver study: (A) Test result w.r.t time sequence. The blue line represents the reference standard force and the red curve are the model predictions. (B) A set of screenshots of the RGB image and their reference standard force and error.} \label{fig1}
\end{figure}

In this paper, we propose a vision-based surgical force prediction model called \textup{R}GB-\textup{P}oint \textup{C}loud \textup{T}emporal \textup{C}onvolutional \textup{N}etwork (\textup{RPC-TCN}). The model is based on a spatial block that encodes information at individual time-steps from a video and a temporal block to reason over sequences of observations. The spatial block combines 2D features (e.g., from an RGB image) and 3D features (e.g., from a 3D point cloud) for a given time, while the temporal block makes use of multiple static features via the Temporal Convolutional Network~\cite{ref_article10} (TCN) to model force change over time. To better abstract the core feature, we apply a pre-trained VGG16 image model~\cite{ref_article11} along with a pre-trained 3D point cloud-based architecture called PointNet~\cite{ref_article12} to extract features from raw visual data and then concatenate these two features to train a TCN time-series model. We evaluate our approach on internally-collected da Vinci surgical video, and show that our model produces highly accurate results. Figure~\ref{fig1} shows representative test result on an \textit{ex vivo} liver.

%Figure1

\subsubsection{Related Work.} Much work has focused on modeling tissue deformation during force prediction, since for reasonably soft material the applied force is positively correlated to the deformation of the tissue surface~\cite{ref_article6, ref_article8, ref_article13}. Therefore, accurately measuring surface deformation in 3D is vitally important for vision-based force estimation. Furthermore, depth data can then be converted to 3D point cloud. The recently proposed PointNet~\cite{ref_article12} directly works on 3D unordered point cloud data, which essentially breaks the pixel order limit of the 2D depth image. The unordered point cloud is robust to camera view points and invariant to transformations, which brings the potential ability to generalize to different objects.

In prior work, Temporal Convolutional Networks (TCN) have been proposed to improve video-based analysis models~\cite{ref_article10}. The input feature vector is the latent encoding of a spatial CNN which corresponds to each frame of the video sequence. Here, we define an observation window which has $n$ frames backward and forward, centered at the current time-step $t$. The label for each window is the force at time $t$, which corresponds to the middle vector in this window. The intuition behind utilizing a time-series model lies with the observation that anatomical surfaces are often deforming continuously. It is then reasonable to introduce time-varying features to determine these forces.

In this paper, our RPC-TCN coalesces the above mentioned features to fully grasp the vision-based properties and then make the force prediction.
%
%
% Methodology
%

\section{Methodology}

\subsection{Dataset Collection}

%Figure2
\begin{figure}[h]
\centerline{\includegraphics[scale=0.375]{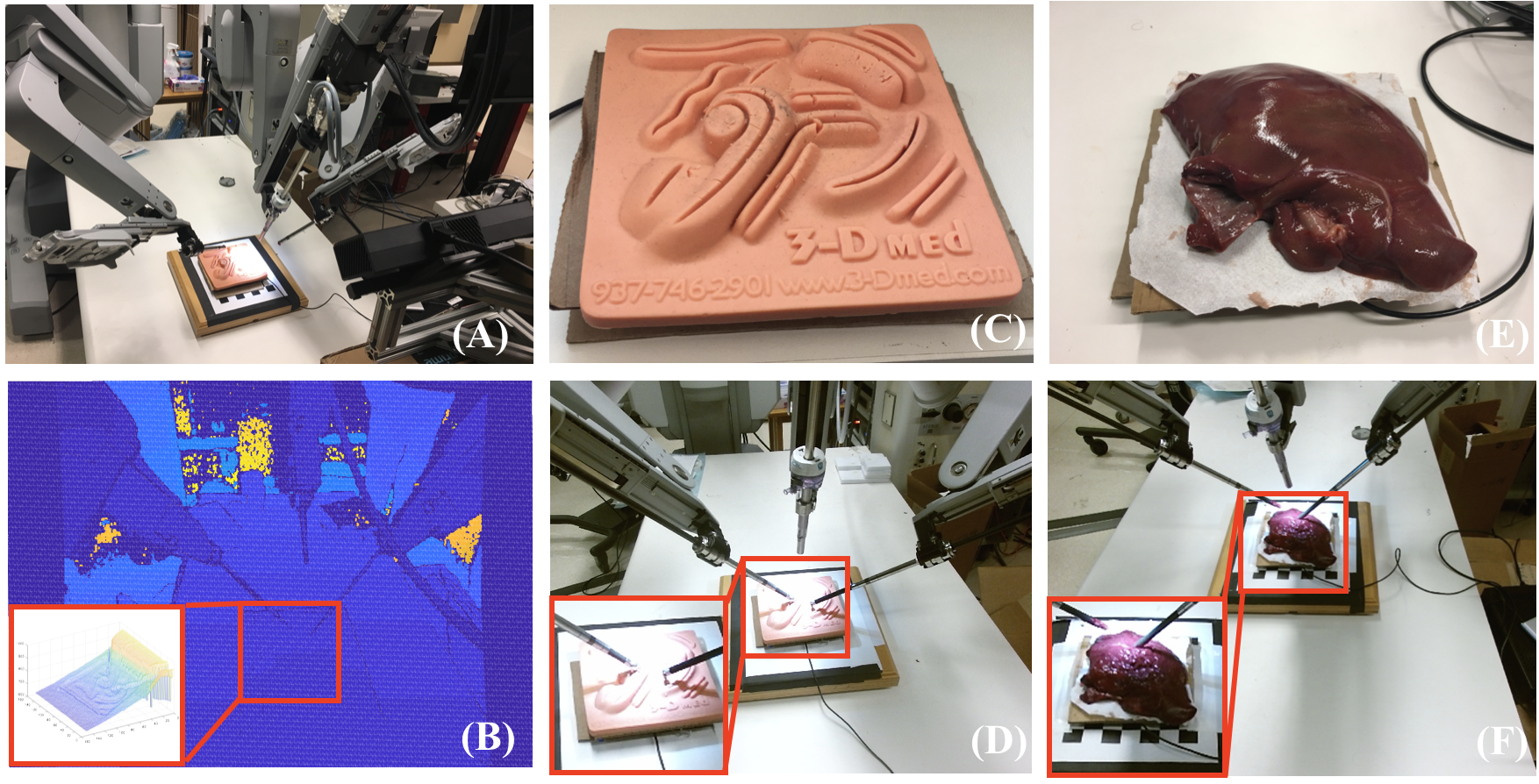}}
\caption{(A) The experimental environment. (B) The depth image from the Kinect2 camera. (C) The phantom used in the experiment. The force sensor is placed under the phantom. (D) RGB image of the phantom. (E) A fresh piece of pig liver. (F) RGB image of the liver.} \label{fig2}
\end{figure}

Since there is no open source dataset for this task, we conduct experiments both in a phantom study and \textit{ex vivo} study to generate our internal dataset. Figure~\ref{fig2} presents the setup details.

To collect force data, we fix the OptoForce 3D Force Sensor underneath the phantom object to record force data. The sensor measures the force a robotic tool applies to the phantom rather than the force at the tool tip itself. The force sensor is accurate up to \SI{12.5e-3}{\newton} and collects 3D force observations (including x, y, and z in the force sensor coordinate). We only use the z-component, which is perpendicular to the planar surface the specimen is placed upon. It will automatically re-bias each time we place an object.

RGB images and depth images are collected using a Kinect2 RGB+D camera. This setup is convenient to demonstrate feasibility of the proposed fusion of RGB images and point clouds for force prediction. In a clinical scenario, a dedicated depth camera is not yet available. However, previous research has validated a learning-based method to estimate dense depth images and surface normal maps from endoscopic surgical video, which results in high-resolution spatial 3D reconstructions to an average error of \SI{0.53}{\mm} to \SI{1.12}{\mm}~\cite{ref_article14}. Based on this result, obtaining 3D point cloud and depth information from endoscopic video is realistically achievable.

The object is fixed in the working area of a standard dual arm da Vinci system~\cite{ref_article1}. As the image stream flows at 30 fps, the RGB data, depth data, and force data are synchronized to be within \SI{10}{\ms}. The Kinect2 RGB+D camera is placed at four different positions to collect multiple points of view to test against model overfitting. 

%Figure3
\begin{figure}
\includegraphics[width=\textwidth]{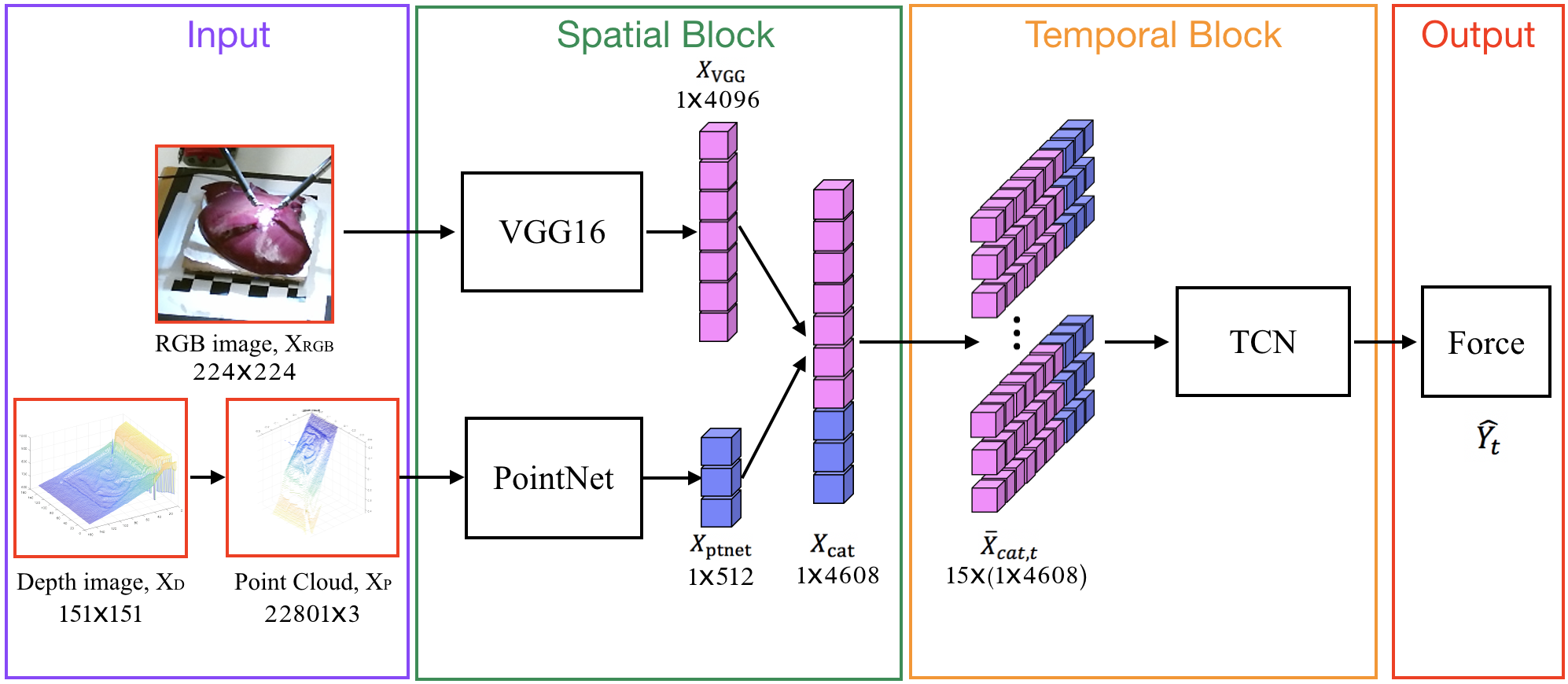}
\caption{The basic architecture of RPC-TCN. The spatial block extracts features from the pre-trained VGG net and PointNet and concatenates two features. The temporal block expands this feature to be 15 frames in a window and predict the force corresponding to the middle frame.} \label{fig3}
\end{figure}

\subsection{RPC-TCN}
\subsubsection{Spatial Block.} Figure~\ref{fig3} shows the overall structure of our RPC-TCN. We use $X_{RGB}\in\mathbb{R}^{224\times224}$ and $X_{D}\in\mathbb{R}^{151\times151}$ to denote RGB image and depth image. The pre-trained VGG16 network has shown good performance at localization and classification tasks~\cite{ref_article11}. In our task, we assume that the movement of the da Vinci tool and the feature change of the phantom is relevant for the force prediction. Thus, we choose to fine tune the pre-trained VGG Network from the ImageNet dataset for later regression. The output feature comes from the 2\textsuperscript{nd} classifier layer, which contains more representative variants than the last layer, such that $X_{VGG}\in\mathbb{R}^{4096}$. Depth image $X_{D}$ is converted to point cloud data in the depth camera's coordinate system. 

\subsubsection{Depth Image to Point Cloud.} In the following formula, ${x}_{D}$ and ${y}_{D}$ refer to the coordinate pixel index in the depth image, and ${z}_{D}$ is its depth value. $\bar{x}_{D}$, $\bar{y}_{D}$, and $\bar{z}_{D}$ refer to the mean values. ${c}_{x}$, ${c}_{y}$ and ${f}_{x}$, ${f}_{y}$ are the intrinsic parameters of principal point and focal length of the depth camera. We use the maximum length in depth image to normalize the point cloud into a unit sphere. The normalized coordinates are

\begin{equation}
x_{pc} = (x_D - c_x)\frac{z_D}{f_x} - \bar{x}_D,
\end{equation}
\begin{equation}
y_{pc} = (y_D - c_y)\frac{z_D}{f_y} - \bar{y}_D, and
\end{equation}
\begin{equation}
z_{pc} = z_D - \bar{z}_D.
\end{equation}

In order to fit the input vector size of the PointNet, the original point cloud is uniformly downsampled from 22801 to 2048 points.  Next, we fine tune the pre-trained PointNet to select the feature from the second-to-last layer, $X_{ptnet}\in\mathbb{R}^{512}$. Finally, we concatenate these two features to a larger one, $X_{cat}=[X_{VGG},X_{ptnet}]\in\mathbb{R}^{4608}$.

%Figure4
\begin{figure}
\includegraphics[width=\textwidth]{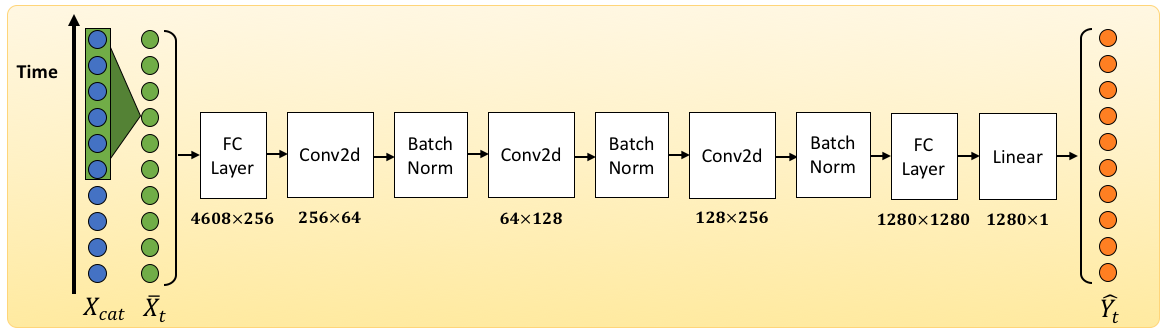}
\caption{Hierarchical structure of the temporal block.} \label{fig4}
\end{figure}

\subsubsection{Temporal Block.} Figure~\ref{fig4} presents the hierarchical structure of the temporal block. Here, we denote the concatenated feature $X_{cat}$ with respect to time as $X_{cat,t}$, then $\bar{X}_{cat,t} = [X_{cat,t-n}, \ldots , X_{cat,t-1}, X_{cat,t}, X_{cat,t+1}, \ldots , X_{cat,t+n}]$. We define the collection of filters in each convolutional layer as $W = \left\{W^{(i)}\right\}_{i=1}^{F_{l}}$ for $W^{(i)}\in\mathbb{R}^{d \times F_{l-1}}$ with a corresponding bias vector $b\in\mathbb{R}^{F_{l}}$, where $l\in\left\{1,\ldots,L\right\}$ is the layer index. Given the signal from the previous layer, $E^{(l-1)}$, we compute activations $E^{(l)}$ with

\begin{equation}
E_{(0)} = f(W*\bar{X}_{cat,t}+b),
\end{equation}

\begin{equation}
E_{(l)} = f(W*\bar{E}_{(l-1)}+b),
\end{equation}

where $f(\cdot)$ is a non-linear activation function and $*$ is the convolution operator. We also perform batch normalization after each convolutional layer. We compare different activation functions and find that the Rectified Linear Units (ReLU) perform best in our experiments. Finally, we use a linear regression at the last fully-connected layer to predict the force, $\hat{Y}_{t}\in\mathbb{R}$. We define $U$ as the filter for the last linear layer $L$ and $c$ as the bias. The process is

\begin{equation}
\hat{Y}_{t} = Linear(UE^{(L)} + c).
\end{equation}

%
%
% Methodology
%
\section{Experiment and Result}

\subsection{Experimental Setting and Dataset}
In total, we obtain 61,473 samples in our phantom study and 44,413 samples in our \textit{ex vivo} liver study. The training data is randomly split to 80\% of the full dataset and 5\% for validation, 15\% for test in both experiments (e.g., the phantom and \textit{ex vivo} were trained and tested separately). The loss function is Mean Squared Error (MSE) and the learning rate is initialized to be \SI{1e-5}  and multiplied by 0.1 every 1000 epochs.

To test the power of the proposed algorithm, we compare to multiple algorithms in both the phantom and \textit{ex vivo} study. We first conduct experiments on traditional single-frame based methods on the RGB images, called Single-frame RGB. In this setup, we use the same VGG16 network to abstract the feature and then construct a convolutional neural network to perform regression. Then we compare the temporal methods, RGB-TCN and Point Cloud-TCN. In these experiments, the features from the spatial block are the same as discussed before, but we test the performance by separately passing them to the same TCN structure. We finally test on the RPC-TCN.

\newcolumntype{P}[1]{>{\centering\arraybackslash}p{#1}}

\begin{table}[!ht]
\caption{Ablation study results.}\label{tab1}
\centering
%\begin{tabular} {p{3cm} p{2cm} p{2cm} p{2cm} p{2cm}}
\begin{tabular}{P{3.3cm}P{2cm}P{2cm}P{2cm}P{2cm}}
\toprule
Algorithm & \multicolumn{2}{c}{Mean Absolute Error (N)} & \multicolumn{2}{c}{Percentage Error}\\
\midrule
& Phantom & \textit{Ex vivo} Liver & Phantom & \textit{Ex vivo} Liver\\
\hline
Single-frame RGB & 7.06 & 10.4 & 3.01\% & 5.45\%\\
RGB-TCN & 2.51 & 1.74 & 1.05\% & 0.913\%\\
Point Cloud-TCN & 2.14 & 1.87 & 0.896\% & 0.983\%\\
 {\bfseries RPC-TCN} &  {\bfseries 1.45} &  {\bfseries 0.814} &  {\bfseries 0.604\% } &  {\bfseries 0.427\%} \\
\bottomrule
\end{tabular}
\end{table}

\subsection{Results and Analysis}
Table 1 displays the prediction accuracy of various algorithms as a comparison on the same dataset. The percentage error is based on the maximum force magnitude in the dataset, which is \SI{-239}{\newton} for the phantom study and \SI{-190}{\newton} for the \textit{ex vivo} liver study. The single-frame RGB is worse than the TCN type methods, which supports the hypothesis that the time-scaled feature is critical to force prediction. Compared to the other two TCN methods, our RPC-TCN presents the best mean absolute error result with \SI{1.45}{\newton}, corresponding to 0.604\% for the phantom study and 0.814 N, corresponding to 0.427\% for the \textit{ex vivo} liver study. 

%Figure5
\begin{figure}
\centerline{\includegraphics[scale=0.35]{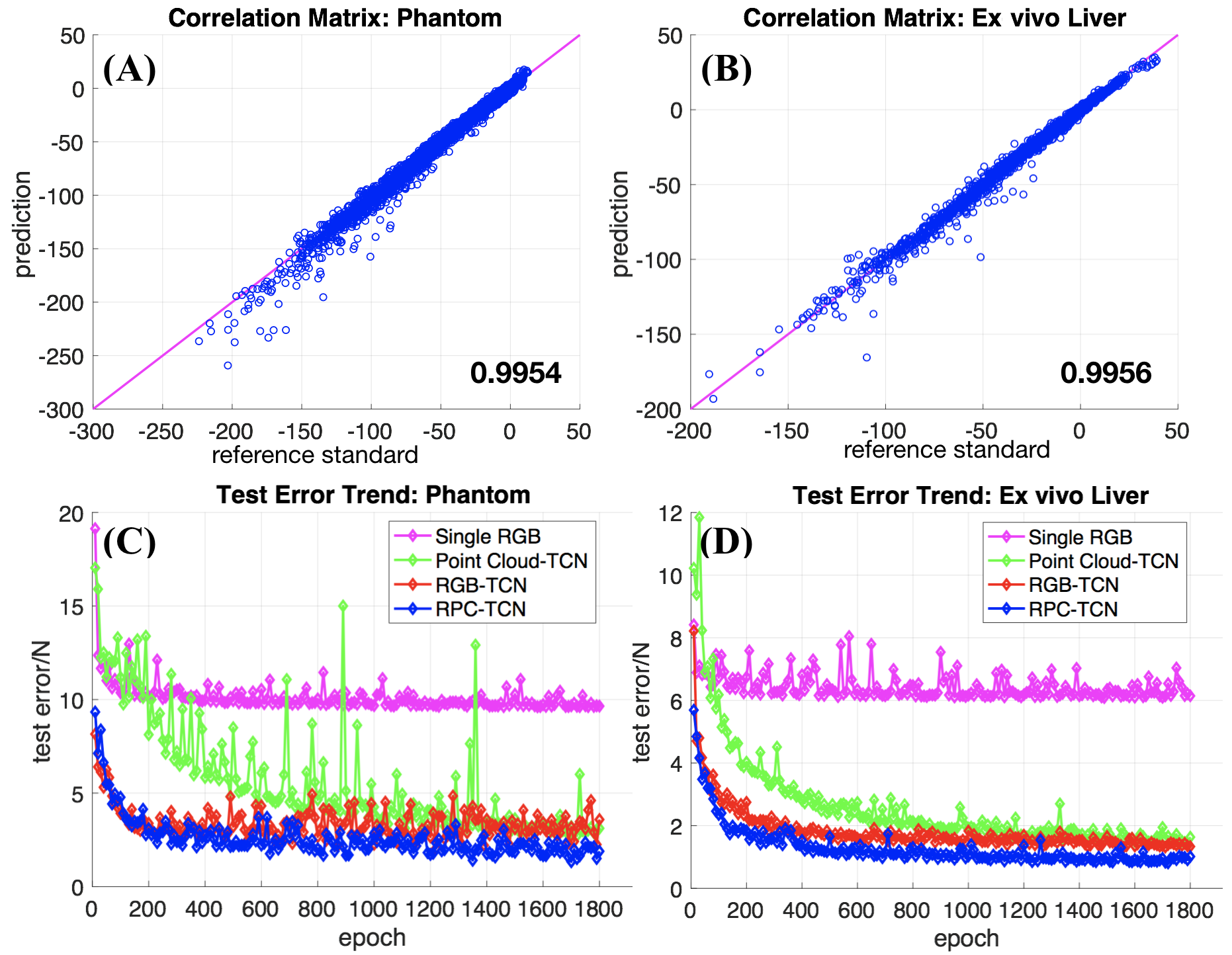}}
\caption{(A) (B) Illustration of the correlation matrix between reference standard force and the prediction force. (C) (D) Test Error trend with training epochs. The error is calculated as mean absolute error of all test data.} \label{fig5}
\end{figure}

Figure 5 (A) (B) displays a correlation plot of our RPC-TCN result. The correlation coefficient is 0.995 and 0.996 for our predictions on the phantom and liver data, respectively, implying a strong relationship between the prediction and the reference standard data. From the test error trend in Figure~\ref{fig5} (C) (D), we find that the single RGB image method presents much higher error, which indicates overfitting. All three TCN based methods can converge to a relatively low test error, while the PC-TCN and the RGB-TCN perform similarly well, but are outperformed by the proposed RPC-TCN suggesting that the use of information from multiple sources is indeed beneficial for vision-based force prediction. 

\begin{figure}
\centerline{\includegraphics[width=12cm]{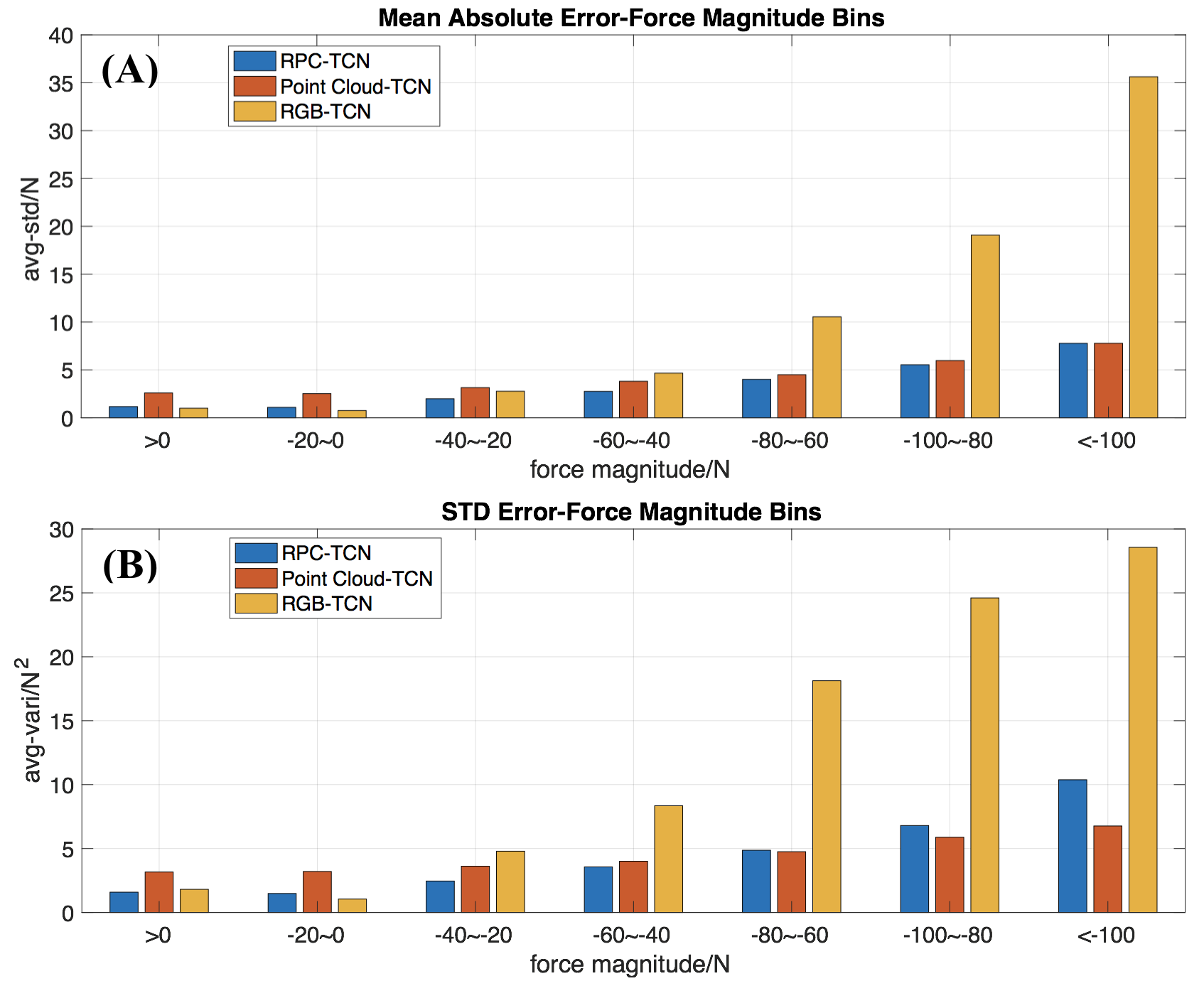}}
\caption{(A) Illustration of mean absolute error distribution w.r.t 7 bins. (B) Illustration of standard deviation error distribution w.r.t 7 bins.} \label{fig6}
\end{figure}

To better understand the error distribution, we divide the force magnitude into 7 bins, each of which spans a \SI{20}{\newton} force interval. Figure~\ref{fig6} (A) (B) show the phantom study result. We calculate the mean absolute error and the standard deviation error in each bin and plot them as comparison. Compared to the Point Cloud-TCN and the RPC-TCN, RGB-TCN shows smaller error in lower force, but it has large error and variation when the force becomes large. This comparison indicates that the RGB-TCN is good at predicting small forces, but it does not perform well for large force prediction.

%Figure6

One of the reasons to this phenomenon is that the training data is biased in force distribution. There are more training samples for smaller forces. A more uniformly-distributed dataset will improve large force prediction. The Point Cloud-TCN shows higher error in low forces, but its prediction error is more uniformly distributed. The reason that Point Cloud-TCN is more steady than RGB-TCN is that features of 3D point cloud are more directly related to deformation than 2D features. This Point Cloud-TCN experiment also shows that only with 3D data, our model achieves good performance, validating the power of depth data. Our proposed RPC-TCN takes advantage of these two information and presents a more consistent error distribution trend regardless of absolute scale.

The \textit{ex vivo} liver study is much closer to a human-organ application compared to the phantom study. Our model still reaches a low error when training and testing in this real organ scenario. We intentionally test large force magnitudes that could cause damage to tissues to be able to predict the onset of excessively large force and, thus, warn clinicians. Training and testing images include specularities, which improves the generalization ability to different surgical scenarios. 

Our current model does not evaluate the transferability to different organs. Current results are reached by training and testing on one single phantom and \textit{ex vivo} liver. We assume the liver properties are similar across different sources, but it still indicates a degree of overfitting to such object. Future study will include testing on multiple organs and considering tissue biomechanical properties. The Kinect camera and OptoForce are convenient tools to demonstrate feasibility of vision-based force estimation. The objects in our experiments are overall flat and thin, which enable the underneath force sensor to measure the applied force change, but the soft tissues are still absorbing part of the touch force. Going forward, however, we will consider setups that are more realistic regarding clinical practice. This includes monocular depth estimation from RGB endoscopic video as in~\cite{ref_article14} and slave-side force sensors to accurately measure tool tip contact force. These must be carefully designed as obtaining ground truth forces \textit{in vivo} is non-trivial.

%
%
% Conclusion
%
\section{Conclusion}

In this paper, we discuss a proof-of-principle system to infer forces during surgical activity from RGB+D video. We propose a convolutional neural network called RGB-Point Cloud TCN (RPC-TCN). This network combines the information from traditional RGB+D images obtained from dense depth imagery, and time series analysis for surgical force prediction in a robotic surgical system. Phantom and \textit{ex vivo} liver experiments yield a mean prediction error to \SI{0.814}{\newton}. Our results on this proof-of-principle prototype are promising and encourage further research on 3D sensing in endoscopy to realize the proposed force sensing approach in clinical practice.

%
%
% Acknowledgement
%
\section*{Acknowledgement}
This work was funded by an Intuitive Surgical Sponsored Research Agreement.

\end{document}